\documentclass[a4paper,twoside]{article}

\usepackage{epsfig}
\usepackage{subcaption}
\usepackage{calc}
\usepackage{amssymb}
\usepackage{amstext}
\usepackage{amsmath}
\usepackage{amsthm}
\usepackage{multicol}
\usepackage{multirow}
\usepackage{apalike}
\usepackage{color}
\usepackage{epstopdf}
\usepackage{graphicx}
\usepackage{xcolor}
\usepackage{amsmath}
\usepackage{multirow}
\usepackage{tikz}
\usepackage{pgfplots}
\usepackage{hyperref}
\usepackage{siunitx}
\usepackage{verbatim}
\usepackage{SCITEPRESS}     
\usepackage{ifthen} 
\usepackage{xfrac}

\definecolor{dimicolor2}{rgb}{0.3, 0.7, 0.1}
\definecolor{kevincolor}{rgb}{1, 0, 0}
\definecolor{dimicolor}{rgb}{0, 0, 0}
\definecolor{newcolor}{rgb}{0, 0, 1}

\definecolor{dimicolor2}{rgb}{0, 0, 0}
\definecolor{kevincolor}{rgb}{0, 0, 0}
\definecolor{dimicolor}{rgb}{0, 0, 0}
\definecolor{newcolor}{rgb}{0, 0, 0}

\definecolor{cameraready}{rgb}{0, 0, 0}

\usepackage[german,textsize=footnotesize,textwidth=1.6cm]{todonotes}
\usepackage{marginnote}

\newboolean{anonymizeIt} 
\setboolean{anonymizeIt}{false} 

\begin{document}
	
	\title{Ground Awareness in Deep Learning for Large Outdoor Point Cloud Segmentation}
	\ifthenelse{\boolean{anonymizeIt}}{

	\author{
		(No name given)
	}
	
  \affiliation{
	  (No affiliation given)}
  }{
		\author{
			\authorname{Kevin Qiu\sup{1}\orcidAuthor{0000-0003-1512-4260}, Dimitri Bulatov\sup{1}\orcidAuthor{0000-0002-0560-2591},
            Dorota Iwaszczuk\sup{2}\orcidAuthor{0000-0002-5969-8533}}
			\affiliation{\sup{1}\scriptsize{Fraunhofer IOSB Ettlingen, Gutleuthausstrasse 1, 76275 Ettlingen, Germany}}
			\affiliation{\sup{2}\scriptsize{Technical University of Darmstadt, Civil and Environmental Engineering Sciences, Darmstadt, Germany}}
	\email{kevin.qiu@iosb.fraunhofer.de}
	}
 }
	
	\abstract{
This paper presents an analysis of utilizing elevation data to aid outdoor point cloud semantic segmentation through existing machine-learning networks in remote sensing, \textcolor{cameraready}{specifically in urban, built-up areas}. 
In dense outdoor point clouds, the receptive field of a machine learning model may be too small to accurately determine the surroundings and context of a point. 
By computing Digital Terrain Models (DTMs) from the point clouds, we extract the relative elevation feature, which is the vertical distance from the terrain to a point. 
RandLA-Net \textcolor{newcolor}{is employed} for efficient semantic segmentation of large-scale point clouds. We assess its performance across three diverse outdoor datasets captured with varying sensor technologies and sensor locations. 
Integration of relative elevation data leads to consistent performance improvements across all three datasets, most notably in the Hessigheim dataset, with an increase of 3.7 \textcolor{cameraready}{percentage points} in average F1 score \textcolor{cameraready}{from 72.35\% to 76.01\%}, by establishing long-range dependencies between ground and objects.
\textcolor{cameraready}{We also} explore additional \textcolor{newcolor}{local} features such as planarity, normal vectors, and 2D features, but their efficacy varied based on the characteristics of the point cloud. 
Ultimately, this study underscores the important role of the \textcolor{newcolor}{non-local} relative elevation feature for semantic segmentation of point clouds in remote sensing applications.
}
		
	\onecolumn \maketitle \normalsize \setcounter{footnote}{0} \vfill
	
\section{Introduction}
Semantic segmentation of outdoor point clouds, \textcolor{dimicolor2}{captured by airborne or terrestrial sensor platforms}, is an important task with use cases in urban planning, mapping, digital twin creation, and vegetation analysis. 
\textcolor{kevincolor}{They also have applications in vehicle navigation, infrastructure management and augmented or virtual reality, where a precise semantic representation of the real environment is required.}
Point clouds, as opposed to 2D orthophotos or 2.5D representations, enhance spatial understanding and allow delineation of vertical structures.
Outdoor point clouds usually cover large areas with a considerable number of points from either LiDAR \textcolor{cameraready}{(Light Detection and Ranging)} sensors or computed through photogrammetry. 
Therefore, segmentation methods need to be fast and efficient. 
{\color{dimicolor}
Semantic segmentation models based on deep learning (DL) became widely used with the advancement of computer processing power and sophisticated architectures, such as PointNet++~\cite{qi2017pointnet++}, KPConv~\cite{thomas2019kpconv}, or RandLA-Net~\cite{hu2020randla}. 
However, these methods may struggle when applied to very large and dense point clouds. 
For example, \textcolor{newcolor}{differentiating} a large horizontal (concrete) roof from an \textcolor{cameraready}{asphalt road} may require \textcolor{cameraready}{a} large receptive field and thus an exaggerated depth \textcolor{cameraready}{and memory footprint} of the neural network, resulting in a high computational cost. 
\textcolor{cameraready}{In point cloud deep learning, the receptive field is characterized by the number of points rather than metric distances. For a point cloud with very high point density, this means that the receptive field may only encompass points within a relatively small spatial area.}

One solution could lie in the integration of \textcolor{newcolor}{additional} features into DL pipelines as surrogates for \textcolor{cameraready}{reliably} large receptive fields.}
\textcolor{dimicolor2}{Fortunately, remote sensing offers a few tools providing long-range context about the vertical dimension of the environment. In particular, elevation over ground is the difference between the \textcolor{kevincolor}{vertical distance} of a point with its closest terrain point and is considered one of the most crucial features for class differentiation. 
Since this closest terrain point can still be arbitrarily far away for large off-terrain objects, \textcolor{kevincolor}{a so-called Digital Terrain Model (DTM) is required, which}} involves evaluating points over large areas, leading to larger receptive fields. 
There are good conventional methods for the computation of DTM, without the need for any labeled points \textcolor{cameraready}{\cite{bulatov2014context,piltz2016volume} while our segmantic segmentation workflow is, of course,} agnostic to the method of DTM derivation.
%
Further local features popular in point cloud analysis, like planarity or point densities and distributions along the $z$\textcolor{newcolor}{-}axis, \textcolor{newcolor}{can also be derived}. \textcolor{newcolor}{In DL-based image processing, it has been shown that adding sensible hand-crafted features improves the overall performance~\cite{audebert2017rgb}.
One goal of the article is to find out which features are useful in the context of point cloud segmentation.}

\textcolor{cameraready}{We argue that RandLA-Net is an efficient network with practical relevance in remote sensing, which can also be easily extended with (hand-crafted) point features.}
Our contributions are as follows:
    1) We extend this network with the relative elevation $h_r$ of a point as well as other point-based 2D and 3D \textcolor{newcolor}{local} features, whereby to calculate $h_r$ for airborne point clouds, we rely on a state-of-the-art point filtering method~\cite{bulatov2014context}.
    2) \textcolor{newcolor}{We demonstrate, on three diverse datasets differing in sensor location, sensor type, landscape type, etc., a consistent improvement of semantic segmentation performance with the addition of $h_r$; in contrast, the additional value of the local features varies depending on the dataset.}
\section{Related Works}
\subsection{Point Cloud Semantic Segmen\-tation \textcolor{newcolor}{U}sing DL}
Convolutional neural networks were extremely successful for the segmentation of 2D images, the reason why  some of the first approaches for 3D data semantic segmentation were projection\textcolor{newcolor}{-}based, like SnapNet~\cite{boulch2018snapnet}. Also inspired by 2D neural networks, 3D convolutions discretize the point cloud into voxels and apply 3D kernels for convolution~\cite{tchapmi2017segcloud}.


The first network to directly operate on the points was the pioneering framework called PointNet~\cite{qi2017pointnet}, using only shared Multi-Layer Perceptrons (MLPs) to process the point features. 
Its successor, PointNet++~\cite{qi2017pointnet++}, learns hierarchical local features through multiple layers and downsampling using Farthest Point Sampling (FPS). PointNeXt~\cite{qian2022pointnext} then improves upon \textcolor{newcolor}{PointNet++} with a better training strategy. Other methods like KPConv~\cite{thomas2019kpconv} use sophisticated learnable kernel-point convolutions.
A downside of all above-mentioned methods is the large processing cost. These networks can either only handle a small number of points at once due to \textcolor{newcolor}{high memory use} or are computationally expensive, for example due to their choice of point sampling method. The fastest sampling method is random sampling, which RandLA-Net~\cite{hu2020randla} employs. With less than a tenth of the parameters as KPConv, for example, it is optimized for high point throughput and is therefore well suited for remote sensing applications. Even though RandLA-Net was published in 2019, no network surpasses it in both speed and performance on the S3DIS 6-fold segmentation task as of the recent publication of PointNeXt, as shown in Table 1 in~\cite{qian2022pointnext}.

Most recently, transformer-based networks like PCT~\cite{guo2021pct} have become popular in the research community. 
The basic transformer architecture calculates global attention between all input tokens, which is immensely computationally expensive. 
Networks like Point Transformer~\cite{zhao2021point} improve on this by only calculating attention between neighboring tokens, similar to SWIN Transformer~\cite{liu2021swin} in the image domain. 
The \textcolor{newcolor}{self-attention mechanisms of transformers open the way} to self-supervised methods, where the model learns inherent features through pretext tasks from the training data without labels. 
For example, Point-MAE~\cite{pang2022masked} and Point-M2AE~\cite{zhang2022point} use point cloud reconstruction as the pretext task. Unfortunately, all transformer methods have a high computational demand and do not yet have a practical relevance \textcolor{newcolor}{in \textcolor{dimicolor2}{large outdoor} point cloud processing}.

\vspace{-2mm}
\subsection{\textcolor{newcolor}{Elevation Data} \textcolor{newcolor}{and Local} Features in \textcolor{newcolor}{S}emantic \textcolor{newcolor}{S}egmentation}

\textcolor{newcolor}{Digital Elevation Models (DEMs) have long been used in object detection.}
In the age of DL in the 2D image domain, the effect of different fusion techniques of RGB and NDSM (Normalized Digital Surface Model, essentially $h_r$ in 2D) in the semantic segmentation of orthophotos was explored in~\cite{qiu2022exploring}, where the addition of NDSM and IR information improved the performance of the U-Net and DeepLabV3+ model;~\cite{audebert2017rgb} come to a similar conclusion.
\textcolor{dimicolor}{In the 3D domain, many experiments with hand-crafted feature sets have been performed. 
Examples are covariance-based features~\cite{maas1999two}, fast point feature histograms~\cite{rusu2009fast}, or signatures of a histogram of orientations~\cite{tombari2010unique}. 
These features can be subject to a conventional classifier, such as Random Forest~\cite{breiman2001random}, whereby e.g.,~\cite{mongus2013detection} have not even learned any thresholds, but have used 3D morphological profiles for detecting buildings in LiDAR point clouds. 
With DL-based approaches, hand-crafted features relying on local neighborhoods have been superfluous, and therefore features exploiting non-local interactions between points gained popularity. 
For example,~\cite{niemeyer2014contextual} test different features, and they find that the height above ground is the most important feature in a \textcolor{newcolor}{Random Forest}-based workflow, where the neighborhood features of a point were also considered. However, the receptive field in a \textcolor{newcolor}{R}andom \textcolor{newcolor}{F}orest approach is typically smaller than in DL, where network depth significantly increases the receptive field.
Furthermore,~\cite{wu2019ground} employ a multi-section plane fitting approach to roughly extract ground points to assist segmentation of objects on the ground, whereby ground filtering has been accomplished implicitly in a weakly-supervised way, such that ground-aware features were utilized with a suitable attention module.
}
The authors of~\cite{YOUSEFHUSSIEN2018191} developed a network based on PointNet that consumes terrain-normalized points along with spectral information. 
In~\cite{9732442}, the ground is detected using a point-based FCN, and a ground aware attention module was added to the segmentation model.
\textcolor{cameraready}{In~\cite{liu2023influence}, RandLA-Net is trained with additional remote sensing features. On a LiDAR dataset colored with RGB information, the addition of a surface normal feature brings a slight improvement in overall mIoU (mean Intersection over Union). The authors of~\cite{pub12335} use RandLA-Net in a hyperspectral urban dataset, where the spectral bands are reduced to an input dimension of 64 using PCA.}
\textcolor{cameraready}{Due to its efficient handling of large point clouds, RandLA-Net is popular in remote sensing \cite{rs14205134,rs15102590,mei2024improving}.}
\vspace{-4mm}
\section{\textcolor{dimicolor}{Methodology}}
\vspace{-2mm}
\subsection{Preliminaries: RandLA-Net}
\vspace{-2mm}

\textcolor{dimicolor2}{According to \cite{qian2022pointnext}, RandLA-Net outperforms competing procedures, such as KPConv~\cite{thomas2019kpconv}, PointNet++~\cite{qi2017pointnet++}, PointNeXt~\cite{qian2022pointnext}, its successor, and Point Transformer~\cite{guo2021pct} regarding the trade-off between accuracy, efficiency, and memory requirements.
Even the most recently published PTv3~\cite{wu2024point} has over 46M parameters, requiring high-end GPUs while RandLA-Net with its 1.3M parameters is able to run on a single Nvidia V100 GPU with only 16 GB of memory, which is our setup. 
Since our point clouds are large and processing needs to be fast and efficient, RandLA-Net is a sensible choice as the network for this paper with practical relevance.}
\par
\textcolor{dimicolor2}{While we refer to~\cite{hu2020randla} for an in-depth understanding of the network architecture, we provide here the most necessary details.}
RandLA-Net uses a U-Net like structure consisting of an encoder and decoder, four layers each, with skip connections. 
In the encoder, the number of points is sampled down and the feature dimension of each point is increased to 512 at the final encoder layer. 
In the decoder, the number of points is increased back to the original point cloud while the feature dimension is reduced to the original eight, but with much higher semantic information. Three fully connected layers, a dropout layer and a softmax follow, resulting in class probabilities for each point. 
The name RandLA-Net originates from the use of \textbf{Rand}om sampling as the downsampling method to discard 75\% of the points after each layer, reducing the input point size by a factor of 256 at the end of the encoder.  
Besides, in each encoder layer, a so-called \textbf{L}ocal feature \textbf{A}ggregation module encodes the relative positions and features of neighboring points of each point using only computationally efficient MLPs. 


\textcolor{newcolor}{This network is able to} process point features additional to the $x, y$ and $z$ coordinates, like RGB values, or those particular features described in the next section. 
The features are first scaled to a dimension size of eight using a fully connected layer and then passed to the encoder, so the additional features barely affect the computational complexity of the network.
\textcolor{newcolor}{The training procedure follows the original implementation with five layers, an input size of 40,000 points, a learning rate of 1e-3 and data augmentation. We use a PyTorch implementation by Idisia Robotics\footnotemark, whereas the original authors use TensorFlow. 
\footnotetext{\href{https://github.com/idsia-robotics/RandLA-Net-pytorch}{\texttt{https://github.com/idsia-robotics/ \\  RandLA-Net-pytorch}}}}

\subsection{\textcolor{newcolor}{Relative Elevation} \textcolor{newcolor}{and Local} \textcolor{newcolor}{F}eatures}
In this section, we describe the \textcolor{newcolor}{additional point} features that are added to RandLA-Net to aid the semantic segmentation task. \textcolor{newcolor}{We compute the relative elevation $h_r$ and other local features that are popular in traditional point cloud processing and remote sensing methods.}

\textcolor{newcolor}{For} $h_r$, we first need to derive the \textcolor{cameraready}{Digital Surface Model (DSM)} and DTM.
The DSM is calculated by sampling the 3D points into a rectangular, equal-spaced 2D grid of a suitable resolution. Each grid cell, or pixel $p$, therefore defines a neighborhood $U(p)$ of 3D points. 
The height value for the DSM of each pixel is determined as the mean $z$-value of the at most $n=4$ highest points within $U$.
This is to avoid random selection of height values at vertical surfaces, like walls, and to reduce noise. 
Because $U(p)$ may be empty, inpainting is done using the heat equation.

The derivation of DTMs is challenging for noisy point clouds with different types of objects. 
\textcolor{newcolor}{There are nowadays efficient DL-based approaches allowing to deal with difficult terrains \cite{gevaert2018deep}, however, they usually need some reference data, which was not available in our case.} 
Thus, we apply a two-step procedure of~\cite{bulatov2014context} to the DSM, \textcolor{newcolor}{which we briefly describe in what follows}. 
First, we identify ground points by considering the circular neighborhood and applying a minimum filter. 
The filter size should have an order of magnitude that corresponds to the size of the largest off-terrain region, for example, the largest building, to ensure that at least one ground point is within the filter. 
If it is too small, points of a building can be spuriously included into the list of ground points, and if it is too large, smaller hills and other elevated regions of the DTM may get lost. 
Here, some too\textcolor{cameraready}{-}large buildings may be interactively masked out. 
In spite of this, this approach may have problems in densely built-up regions and also when there is a high percentage of outliers in the elevation map.
As a consequence, the functional that approximates the ground points should be robust against outliers in the data.
We chose the 2.5D cubic spline surface 
computed by minimization of a functional as in~\cite{BL1} in the $L_1$-norm:  
\begin{eqnarray}
\begin{array}{c}
L(\mathbf{z})=\displaystyle {(1-\lambda) \sum_{m=1}^M \left|z\left(x_m, y_m\right) - z_m \right|+} 
\\
\displaystyle {+\lambda \int_{x,y} \left ( \left|z_{xx}\right|+2\left|z_{xy} \right|+\left|z_{yy}\right| \right) \mbox{d}x\,\mbox{d}y} +
\\
\displaystyle {\varepsilon \sum_{\mbox{nodes}} \left (\left |z_x \right| +\left | z_y \right| \right )}, 
\end{array}
\label{eq_l1}
\end{eqnarray}
\noindent where $\{(x_m, y_m, z_m)\}^M_{m=1}$ are the coordinates of the ground points, $\mathbf z = z(x, y)$ is the function value of the point $(x, y)$ for a Sibson-element of a cubic spline, $\lambda \approx 0.7$ is a data fidelity parameter, and $\varepsilon$ is a small positive scalar\textcolor{newcolor}{, supposed to guarantee numerical stability}. 
\textcolor{newcolor}{The DTM value for each pixel $p$ is given by the spline parameters.}
The difference between the $z$ coordinate of a point and the DTM of the corresponding pixel yields its relative elevation $h_r$.

Using the same neighborhood $U$, the number of points per pixel $\nu$ and the variance of their elevations $\sigma(z)$ are determined. 
These \textcolor{newcolor}{local} features are calculated in the 2D plane and assigned to every point in $U$. 

\begin{figure*}[h]
    \centering
    \begin{subfigure}{0.3\textwidth}
        \centering
        \includegraphics[width=\textwidth]{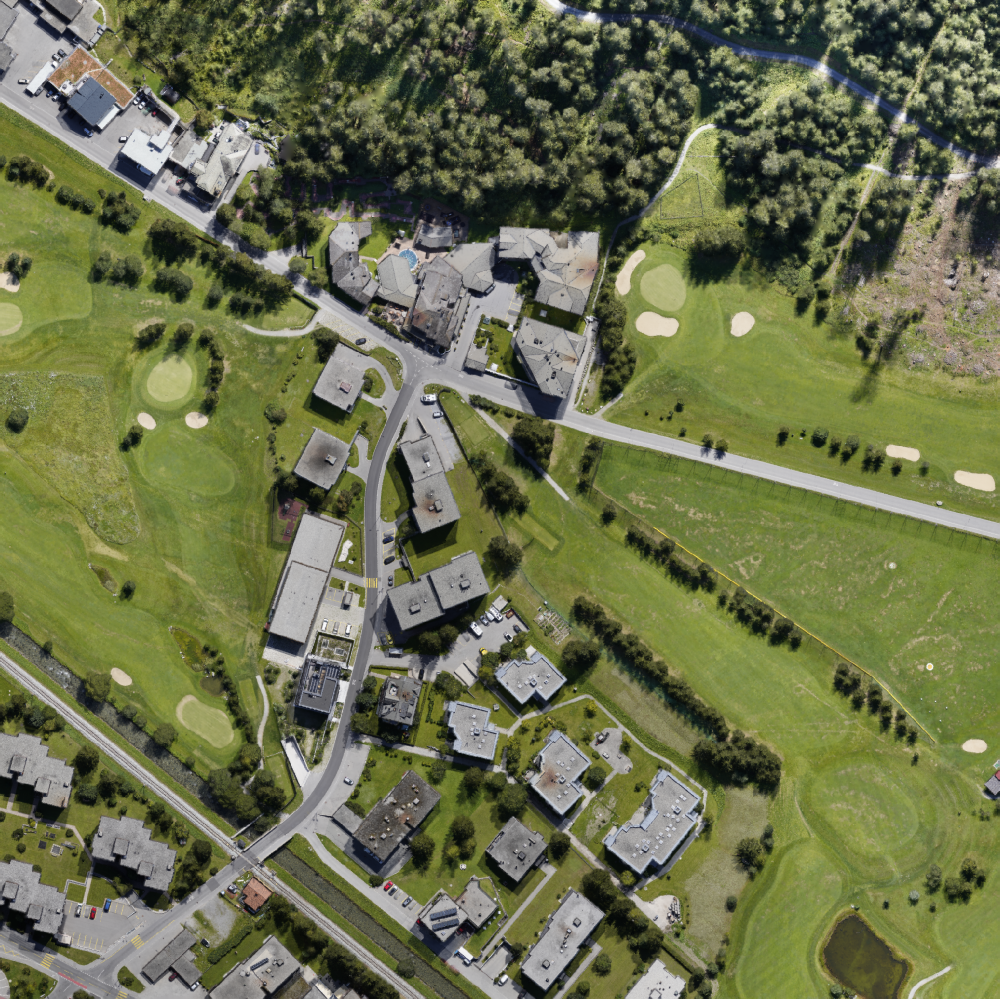}
        \caption{Orthoprojection}
        \label{fig:davos_ortho}    
    \end{subfigure}
    \begin{subfigure}{0.3\textwidth}
        \centering
        \includegraphics[width=\textwidth]{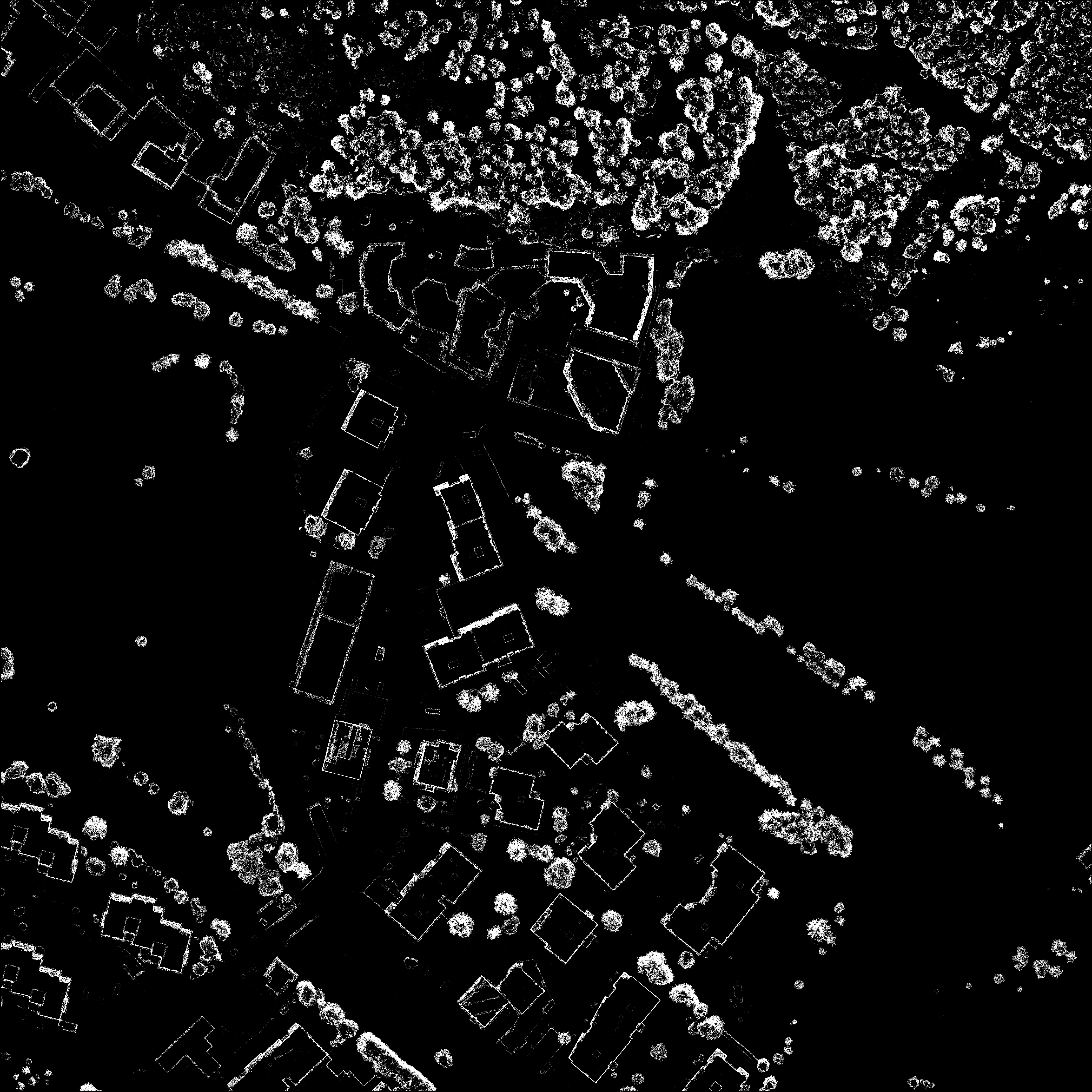}
        \caption{$\nu$}
        \label{fig:davos_numpix}
    \end{subfigure}
    \begin{subfigure}{0.3\textwidth}
        \centering
        \includegraphics[width=\textwidth]{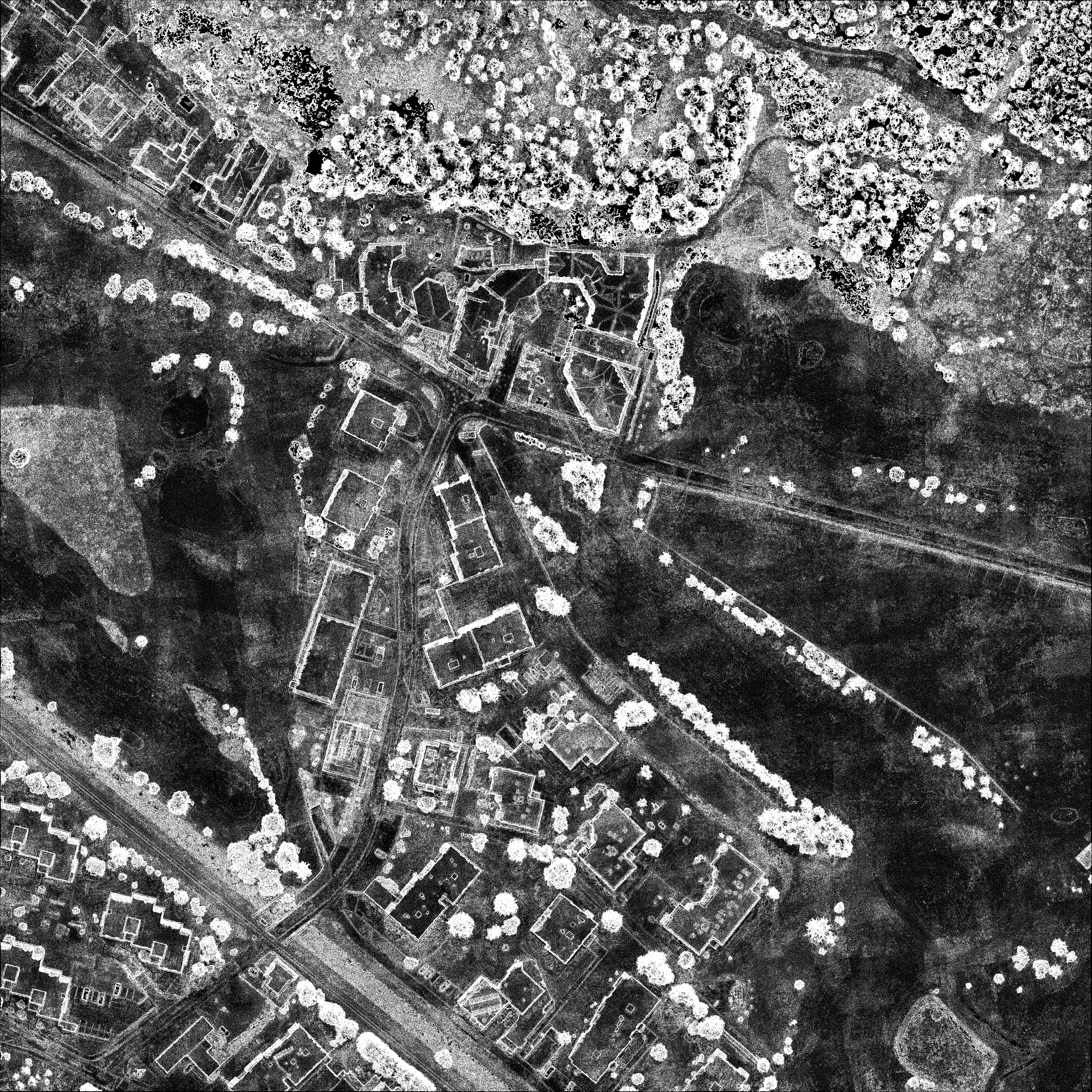}
        \caption{$\sigma(z)$}
        \label{fig:davos_varz}
    \end{subfigure}
    
    \begin{subfigure}{0.3\textwidth}
        \vspace{0.25cm}
        \centering
        \includegraphics[width=\textwidth]{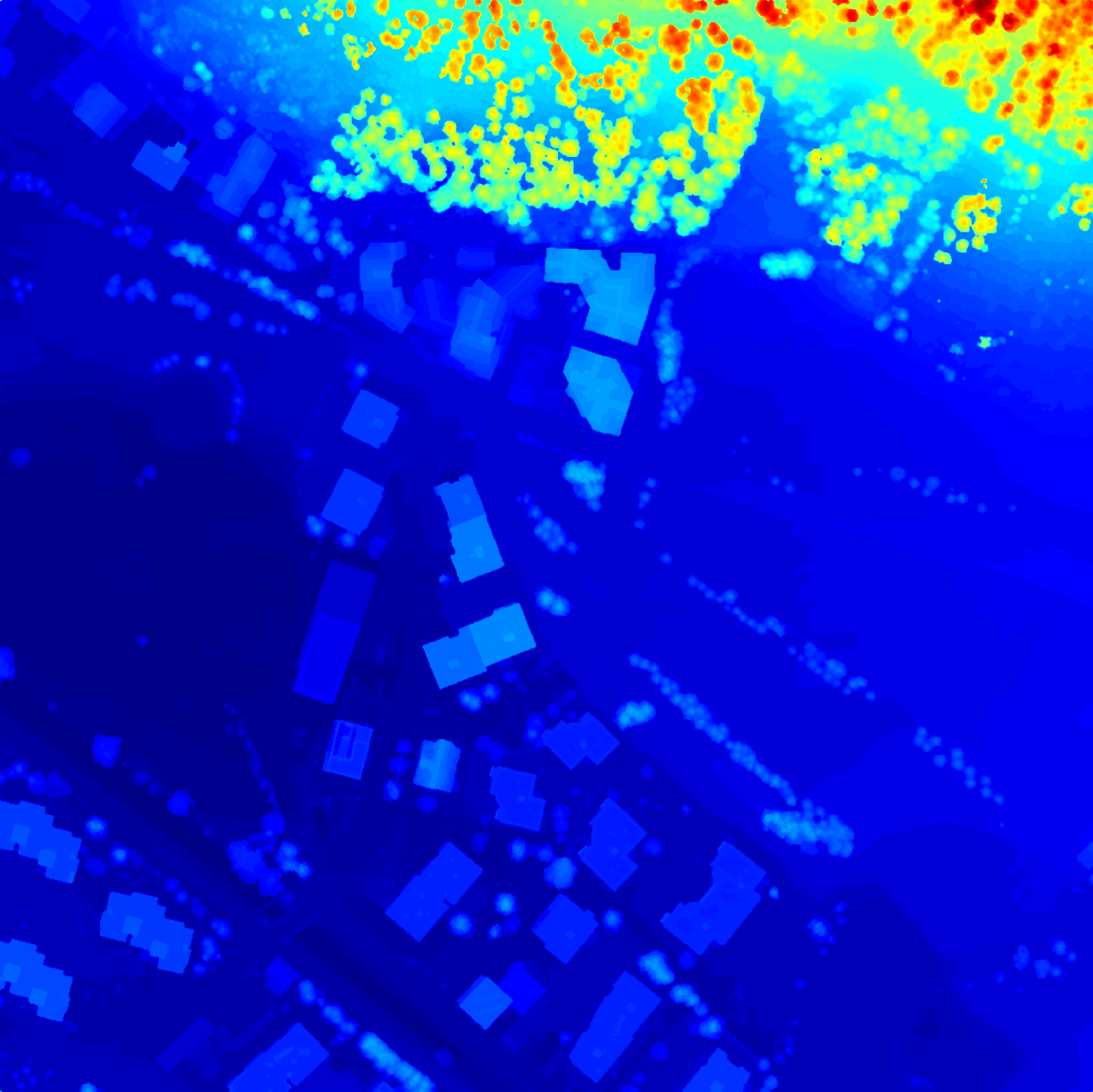}
        \caption{DSM}
        \label{fig:davos_dsm}
        \vspace{0.25cm}
    \end{subfigure}
    \begin{subfigure}{0.3\textwidth}
        \centering
        \includegraphics[width=\textwidth]{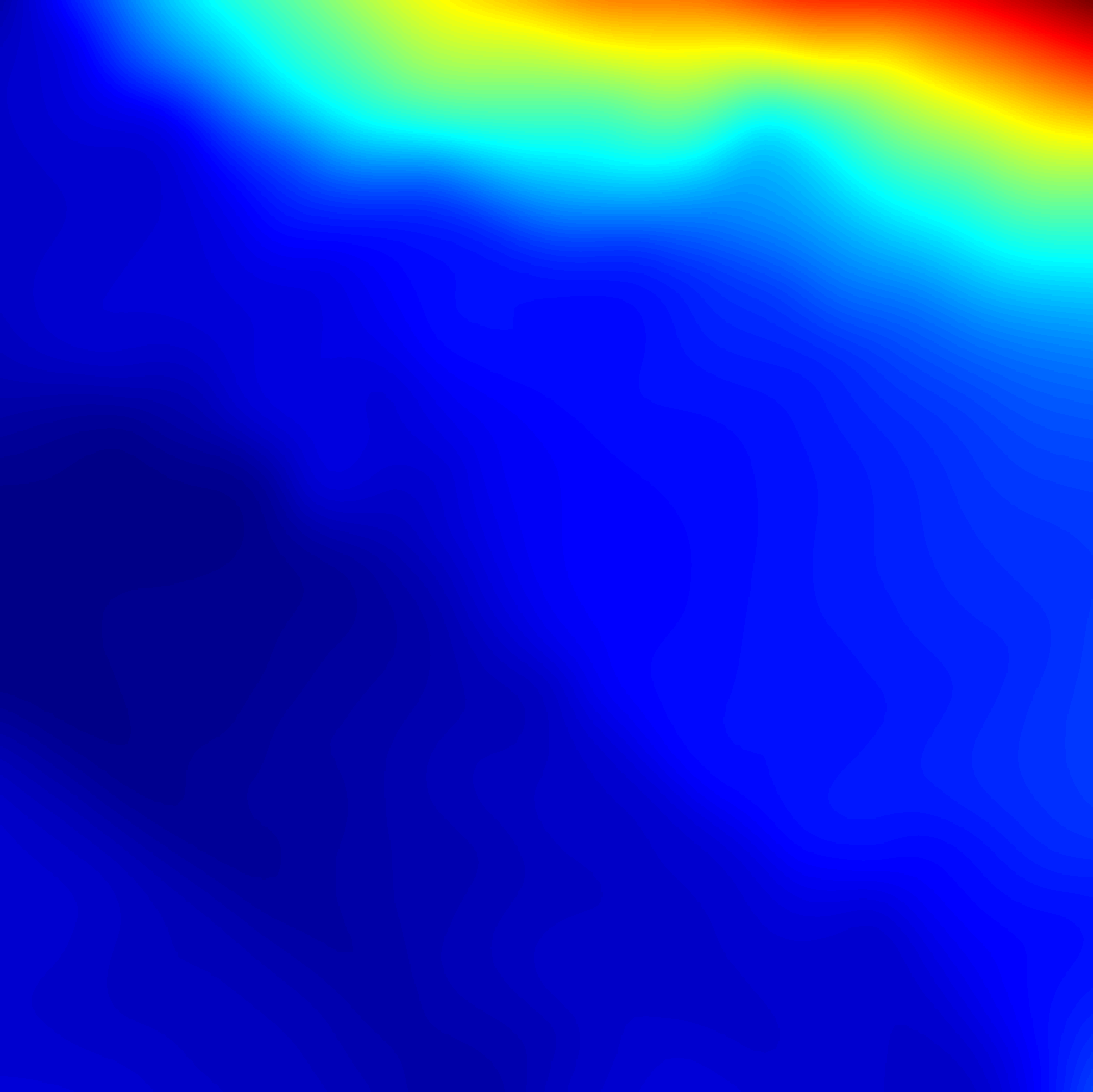}
        \caption{DTM}
        \label{fig:davos_dtm} 
        \vspace{0.25cm}
    \end{subfigure}
    \begin{subfigure}{0.3\textwidth}
        \centering
        \includegraphics[width=\textwidth]{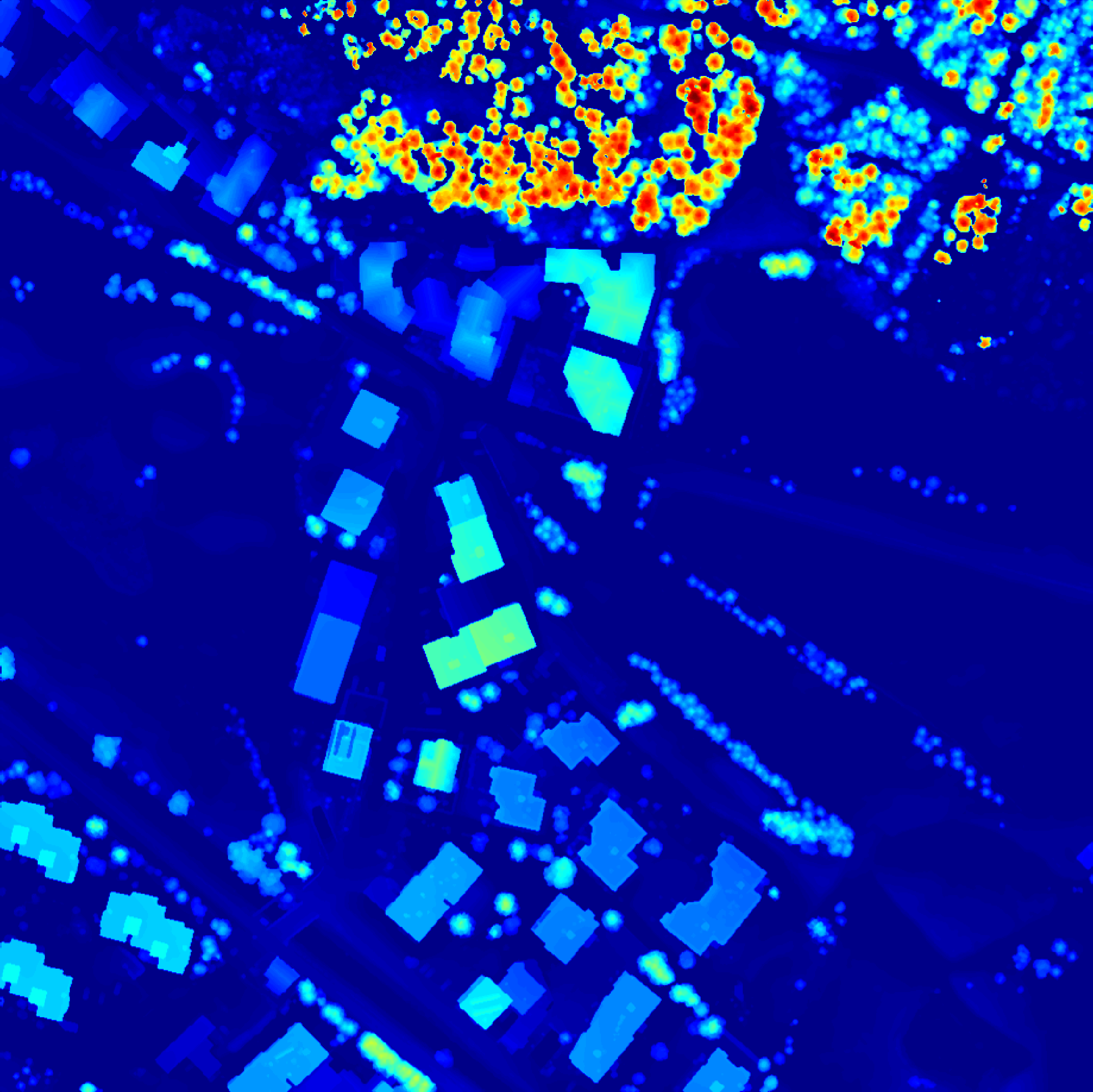}
        \caption{Normalized DSM}
        \label{fig:davos_ndsm}
        \vspace{0.25cm}
    \end{subfigure}

    \caption{View of the orthoprojection, DSM, DTM, NDSM, and 2D features \textcolor{cameraready}{(number of points $\nu$ and variance of the points $\sigma(z)$ in a “pixel” of an 2D grid)} of the fifth Davos tile of the Swiss2DCities dataset. The Normalized DSM, or NDSM, is two-dimensional and calculated by subtracting the DTM from the DSM.}
    \label{fig:davos_relz_calculation}
\end{figure*}

{\color{dimicolor}Planarity $\eta$ and normals $n_z$ are derived directly from the point cloud. 
\textcolor{cameraready}{From $K$ nearest neighbors of every point}, we compute the structure tensor. 
Its eigenvalues $\lambda_1, \lambda_2, \lambda_3$, sorted in descending order, and the corresponding eigenvectors $v_1, v_2, v_3$ are used to calculate 
\begin{eqnarray}
    \eta = \frac{\lambda_2 - \lambda_3}{\lambda_1}, n_z = (v_3)_z,\label{eq_eigen}
\end{eqnarray}
as well as a few other measures (omnivariance, linearity, etc.) mentioned in the implementation of~\cite{WEINMANN2015286}. 
We chose $\eta$ and $n_z$ in \eqref{eq_eigen} because planarity turned out to be the most distinctive measure in many publications \cite{weinmann2017geometric}, among others, while the normal vector direction is not directly coupled to the eigenvalues. 
Note that we decided to make this feature invariant with respect to rotations around the $xy$-plane and modulo $\pi$ around $z$-axis.}
All \textcolor{dimicolor}{3D} features are calculated on the full point clouds without any downsampling such as grid sampling. Due to the extreme irregularity of the point cloud in the terrestrial Toronto3D dataset (see next section), especially in regions further away from the sensor path, we decided not to use our two-step method for DTM computation. Instead, a simple RANSAC algorithm was used to find the ground plane of each of the four tiles. The resulting elevation data is not as accurate as in the other datasets, since the ground is not perfectly flat. 

Figure~\ref{fig:davos_relz_calculation} shows the 2D features $\nu$ and $\sigma(z)$ as well as the steps needed to calculate the relative elevation information on a tile of the Swiss3DCities dataset. 
The number-of-points feature in Fig.~\ref{fig:davos_numpix}, for example, has high values at building walls and tall vegetation, where many points are stacked in the \textcolor{cameraready}{$z$-direction}.
Table~\ref{tab:features} shows all features that are used in this paper to aid semantic segmentation.

\begin{table}[]\small
\centering
\caption{Description of the point-wise input features for RandLA-Net that are utilized in this paper. \textcolor{newcolor}{Abbreviations: NNs = Nearest neighbors, NV = Normal Vector, Feat. = Feature, ?D = Dimensionality.}}
\label{tab:features}
\begin{tabular}{llc}\hline
\textbf{Feat.} & \textbf{Description}               & \textbf{?D} \\ \hline
color            & RGB information                    & 3D             \\
$h_r$             & Rel. height above the terrain & 3D             \\
$\eta$ & \begin{tabular}[c]{@{}l@{}}Planarity calculated using cova-  \\ riance matrix of 10 NNs\end{tabular}            & 3D \\
$n_z$  & \begin{tabular}[c]{@{}l@{}}Absolute value of NV$_z$  \\ computed over 100 NNs \end{tabular} & 3D \\
$\sigma(z)$      & Variance of the points in $U$ & 2D             \\
$\nu$            & Number of points in $U$       & 2D             \\ \hline
\end{tabular}%
\end{table}

\section{Datasets}
\vspace{-2mm}
A diverse selection of datasets, listed in  Table~\ref{tab:datasets}, were chosen for comprehensive analysis. \textcolor{cameraready}{Different c}apturing techniques, e.g.~LiDAR vs. Photogrammetry, and sensor location, e.g.~ aerial or terrestrial, yield very different point configurations. 
\begin{table*}[]
\centering
\caption{Comparison of the three outdoor point cloud datasets used in this paper.}
\label{tab:datasets}
\begin{tabular}{ccccccc}
\hline
\textbf{Dataset} &
  \textbf{Platform} &
  \textbf{Sensor} &
  \textbf{\#Points} &
  \textbf{Classes} &
  \textbf{Year} \\ \hline
Swiss3DCities  &  Air- & Photogram- & \multirow{2}{*}{67.7M}  & \multirow{2}{*}{5} & \multirow{2}{*}{2020} \\
Davos          &  borne & metric    &         &  &\\
\hline
Hessigheim & Air- &   Riegl VUX-1LR   &  \multirow{2}{*}{125.7M} & \multirow{2}{*}{11} & \multirow{2}{*}{2021} \\
 March 2018  &  borne & LiDAR    &         &  &\\
\hline
Toronto3D & Vehicle &   Teledyne Optech   &  \multirow{2}{*}{78.3M}  & \multirow{2}{*}{8} & \multirow{2}{*}{2020} \\
 March 2018  &  MMLS & Maverick LiDAR    &         &  &\\
\hline  

\end{tabular}%
\end{table*}

Swiss3DCities~\cite{can2021semantic} is an UAV-based dataset, covering a total area of \SI{2.7}{\kilo\meter\squared} from three different Swiss cities with a ground sampling distance (GSD) of \SI{1.28}{\centi\meter}. The high-resolution point clouds were derived photogrammetrically and labeled into five classes. 
The authors emphasize uniform density and completeness, including on vertical surfaces, through oblique captures. 
We use the “medium” density point clouds and only the five tiles from the city of Davos to speed up the training time.
For validation, we use the fifth Davos tile, which consists of about 12.5M points.

The Hessigheim dataset~\cite{KOLLE2021100001} is an airborne RGB LiDAR dataset of the city Hessigheim in Germany, with a GSD of \SIrange{2}{3}{\centi\meter}. The point density is about 800 points per square meter, but is quite heterogeneous along the vertical dimension. 
We chose the most popular March 2018 capture \textcolor{newcolor}{and calculated the local features using the same parameters as in Swiss3DCities.}
The dataset consists of a train, validation, and test set, but the labels for the latter are not publicly available, requiring submission for evaluation of the approximately 52M points.

The Toronto3D dataset~\cite{tan2020toronto} is also captured using LiDAR. 
However, the sensor was mounted on a driving vehicle, resulting in large density variations due to vastly different distances from the car to the scene. 
The LiDAR and RGB camera scans along a \SI{1}{\kilo\meter} stretch of road are merged, resulting in even more density variation due to varying speed of the vehicle. 
The authors deliberately kept all points of the entire approximately \SI{100}{\meter} scan radius to mimic real life point clouds.
The dataset is divided into four tiles, with the second tile used as validation, as was instructed by \cite{tan2020toronto}.
\vspace{-2mm}
\section{Results}

\begin{figure*}[h]
    \centering
    \begin{subfigure}{0.45\textwidth}
        \centering
        \includegraphics[width=\textwidth, trim=0cm 0cm 0cm 4cm, clip]{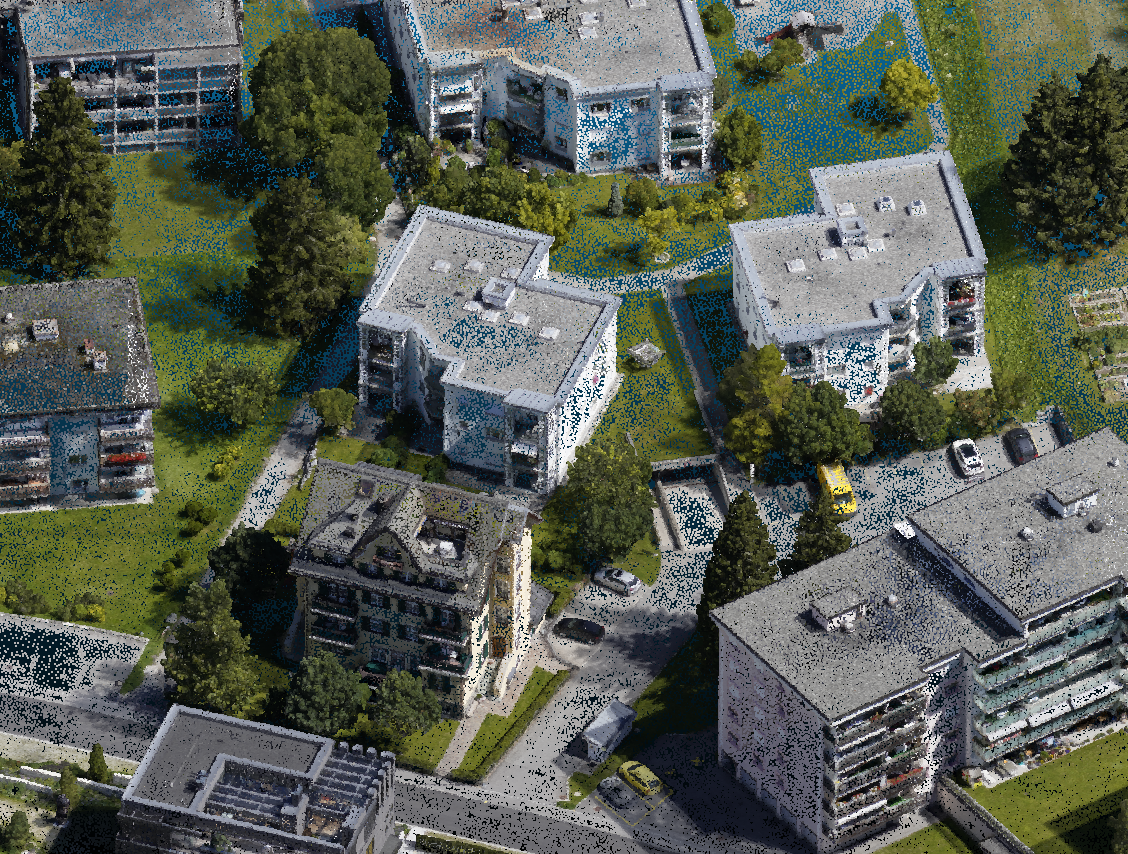}
        \caption{RGB}
        \label{fig:davos_rgb}
    \end{subfigure}
    \hfill
    \begin{subfigure}{0.45\textwidth}
        \centering
        \includegraphics[width=\textwidth, trim=0cm 0cm 0cm 4cm, clip]{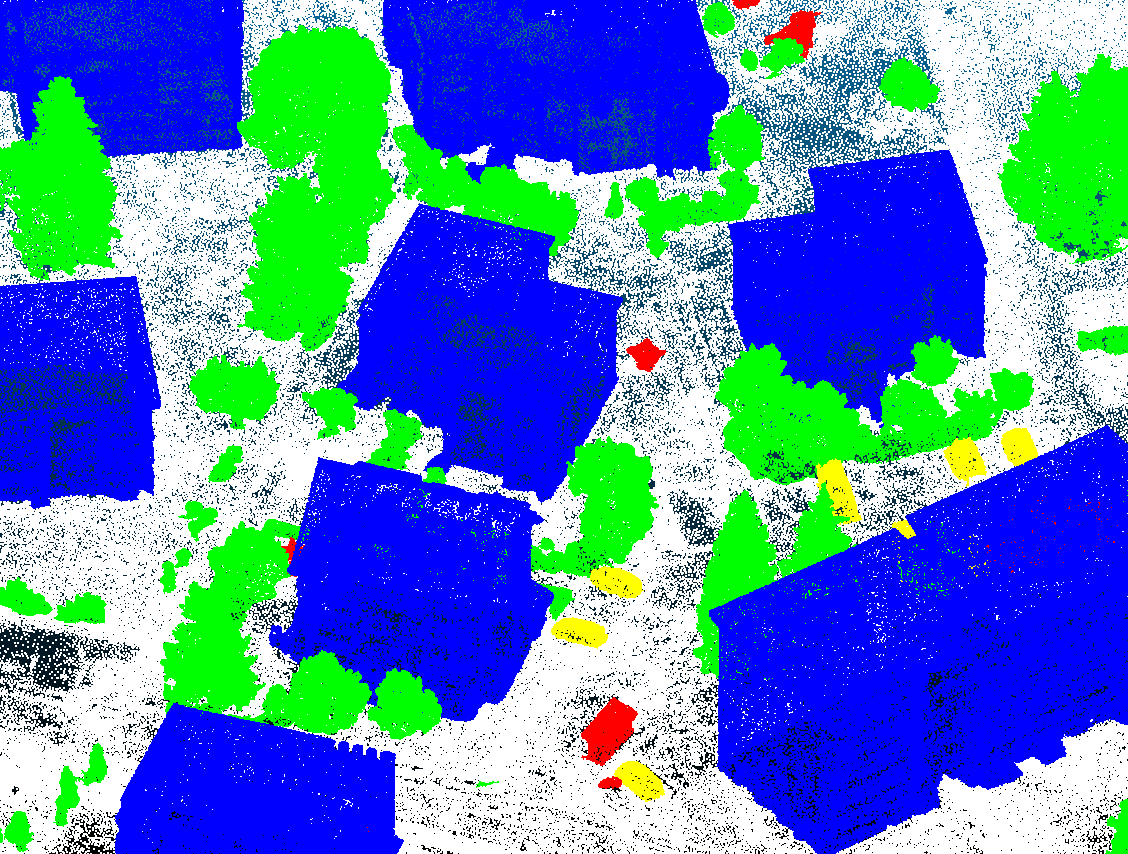}
        \caption{Ground truth}
        \label{fig:davos_gt}
    \end{subfigure}
    \begin{subfigure}{0.45\textwidth}
        \vspace{0.25cm}
        \centering
        \includegraphics[width=\textwidth, trim=0cm 0cm 0cm 4cm, clip]{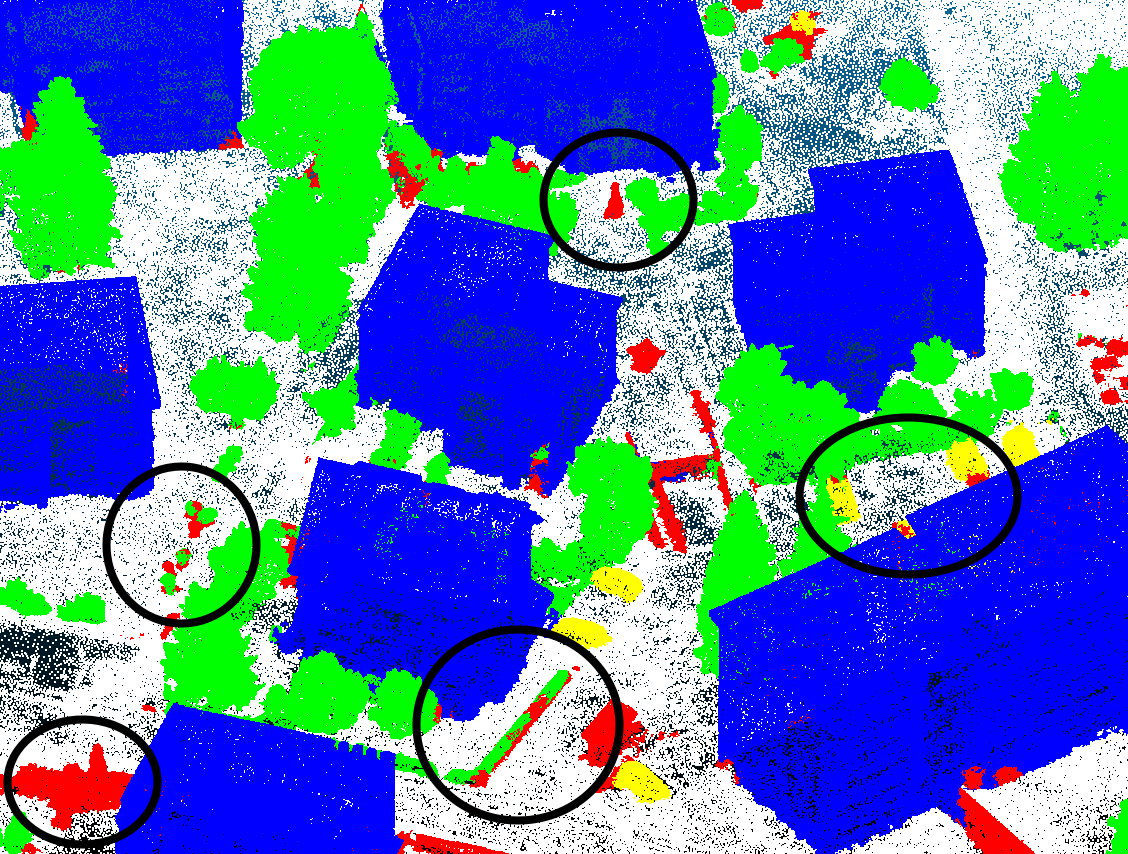}
        \caption{Prediction ``color''}
        \label{fig:davos_color}
        \vspace{0.25cm}
    \end{subfigure}
    \hfill
    \begin{subfigure}{0.45\textwidth}
        \vspace{0.25cm}
        \centering
        \includegraphics[width=\textwidth, trim=0cm 0cm 0cm 4cm, clip]{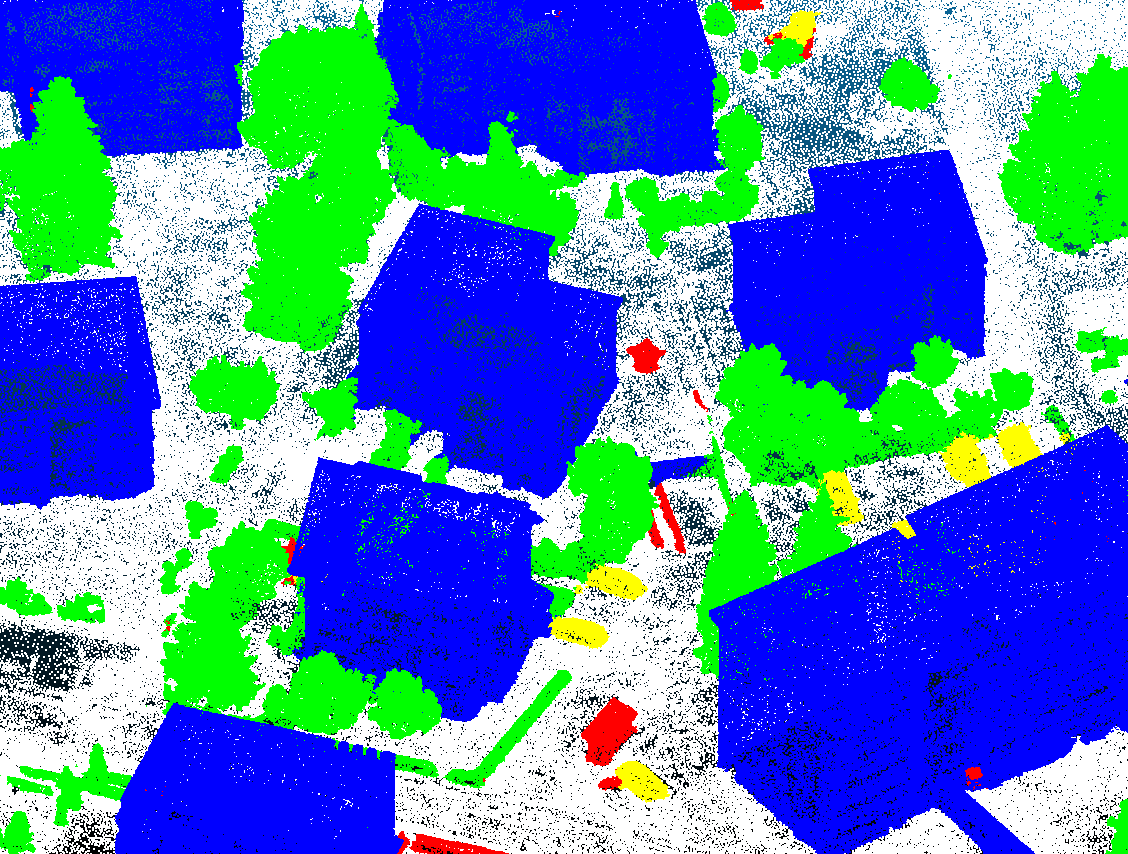}
        \caption{Prediction ``all''}
        \label{fig:davos_all}
        \vspace{0.25cm}
    \end{subfigure}
    \caption{\textcolor{newcolor}{The results on the fifth Davos validation tile of the Swiss3DCities dataset, with predictions of two configurations and corresponding ground truth. Some errors in Fig.~\ref{fig:davos_color} that are not present in Fig.~\ref{fig:davos_all} are circled in black.}}
    \label{fig:davos}
\end{figure*}

In the Davos subset of the Swiss3DCities dataset, as shown in Table~\ref{tab:swiss3d}, the addition of the relative information $h_r$ feature \textcolor{newcolor}{significantly} improve\textcolor{newcolor}{s} the mIoU from 66.70\% with just color features to 69.81\%. This jump is larger than after the addition of color features to the configuration \textcolor{cameraready}{containing point coordinates only}. 
Adding \textcolor{newcolor}{elevation and all local} features improves the results even more, to an mIoU of 71.20\%. The largest jump in IoU score can be seen in the small urban asset class, with the terrain class not far behind.
Figure~\ref{fig:davos} shows the qualitative results along with the RGB point cloud and ground truth. The ``all'' configuration, using all features listed in Tab~\ref{tab:features}, in Fig.~\ref{fig:davos_all} shows fewer artifacts, mostly of the urban asset class (red), than the “color” configuration in Fig.~\ref{fig:davos_color}, making most other classes like building (blue) and vegetation (green) more accurate.
It should be noted that the ground truth data provided may not be entirely accurate or comprehensive\textcolor{cameraready}{, partially due to inherent ambiguities}. For instance, certain objects such as hedges, man-made structures like walls or entrances to underground parking facilities have been mislabeled as terrain.  \textcolor{kevincolor}{These areas are however predicted as either urban asset or building, like in the bottom right corner of the Figures.}

\begin{table*}[t!]
\caption{Performance of RandLA-Net with different feature configurations on the Davos subset validation tile of the Swiss3DCities dataset. \textcolor{newcolor}{The first configuration only uses the \textcolor{cameraready}{$xyz$} point coordinates}.}
\label{tab:swiss3d}
\centering
\begin{tabular}{llllllll}
\hline
\textbf{Features} &
  \textbf{\begin{tabular}[c]{@{}l@{}}OA\\
  \end{tabular}} &
  \textbf{\begin{tabular}[c]{@{}l@{}}Mean\\ IoU\end{tabular}} &
  \textbf{\begin{tabular}[c]{@{}l@{}}IoU\\ terrain\end{tabular}} &
  \textbf{\begin{tabular}[c]{@{}l@{}}IoU\\ constr.\end{tabular}} &
  \textbf{\begin{tabular}[c]{@{}l@{}}IoU\\ urban a.\end{tabular}} &
  \textbf{\begin{tabular}[c]{@{}l@{}}IoU\\ veget.\end{tabular}} &
  \textbf{\begin{tabular}[c]{@{}l@{}}IoU\\ vehicle\end{tabular}} \\ \hline
-            & 91.77 & 66.23 & 86.54 & 81.55 & 17.24 & 92.43 & 53.51 \\
color        & 91.31 & 66.70 & 84.76 & 82.17 & 16.49 & 92.17 & 57.91 \\
color+$h_r$ & 93.08 & 69.81 & 87.95 & 83.17 & 25.30 & \textbf{94.13} & \textbf{58.50} \\
all          & \textbf{93.31} & \textbf{71.20} & \textbf{88.06} & \textbf{85.19} & \textbf{32.30} & 93.40 & 57.15 \\ \hline
\end{tabular}
\end{table*}

\begin{table*}[ht]
\caption{Performance of RandLA-Net with different input features on the Hessigheim March 2018 test set.}
\label{tab:hessigheim}
\resizebox{\textwidth}{!}{%
\begin{tabular}{llllllllllllll}
\hline
\textbf{Feat.} &
  \textbf{OA} &
  \textbf{mF1} &
  \textbf{\begin{tabular}[c]{@{}l@{}}F1 L.\\ Veg.\end{tabular}} &
  \textbf{\begin{tabular}[c]{@{}l@{}}F1 I.\\ Surf.\end{tabular}} &
  \textbf{\begin{tabular}[c]{@{}l@{}}F1 \\ Car\end{tabular}} &
  \textbf{\begin{tabular}[c]{@{}l@{}}F1 U.\\ Furn.\end{tabular}} &
  \textbf{\begin{tabular}[c]{@{}l@{}}F1 \\ Roof\end{tabular}} &
  \textbf{\begin{tabular}[c]{@{}l@{}}F1\\ Fac.\end{tabular}} &
  \textbf{\begin{tabular}[c]{@{}l@{}}F1\\ Shr.\end{tabular}} &
  \textbf{\begin{tabular}[c]{@{}l@{}}F1\\ Tree\end{tabular}} &
  \textbf{\begin{tabular}[c]{@{}l@{}}F1\\ Grav.\end{tabular}} &
  \textbf{\begin{tabular}[c]{@{}l@{}}F1 V.\\ Surf.\end{tabular}} &
  \textbf{\begin{tabular}[c]{@{}l@{}}F1\\ Chi.\end{tabular}} \\ \hline
color        & 86.00 & 72.35 & 91.53 & 86.88 & \textbf{61.78} & \textbf{50.64} & 94.72 & 79.07 & 56.75 & 96.45 & 30.06 & 69.30 & 88.68 \\
color+$h_r$ & \textbf{87.67} & \textbf{76.01} & \textbf{92.40} & \textbf{87.52} & 55.61 & 49.63 & \textbf{96.52} & 79.09 & \textbf{64.84} & \textbf{96.76} & \textbf{45.86} & \textbf{78.47} & \textbf{89.42} \\
all          & 85.24 & 74.62 & 90.17 & 84.11 & 60.56 & 49.74 & 94.63 & \textbf{79.57} & 63.93 & 95.70 & 40.67 & 72.34 & 89.42 \\ \hline
\end{tabular}%
}
\end{table*}

\begin{figure*}[h]
    \centering
    \begin{subfigure}{0.3\textwidth}
        \centering
        \includegraphics[width=\textwidth]{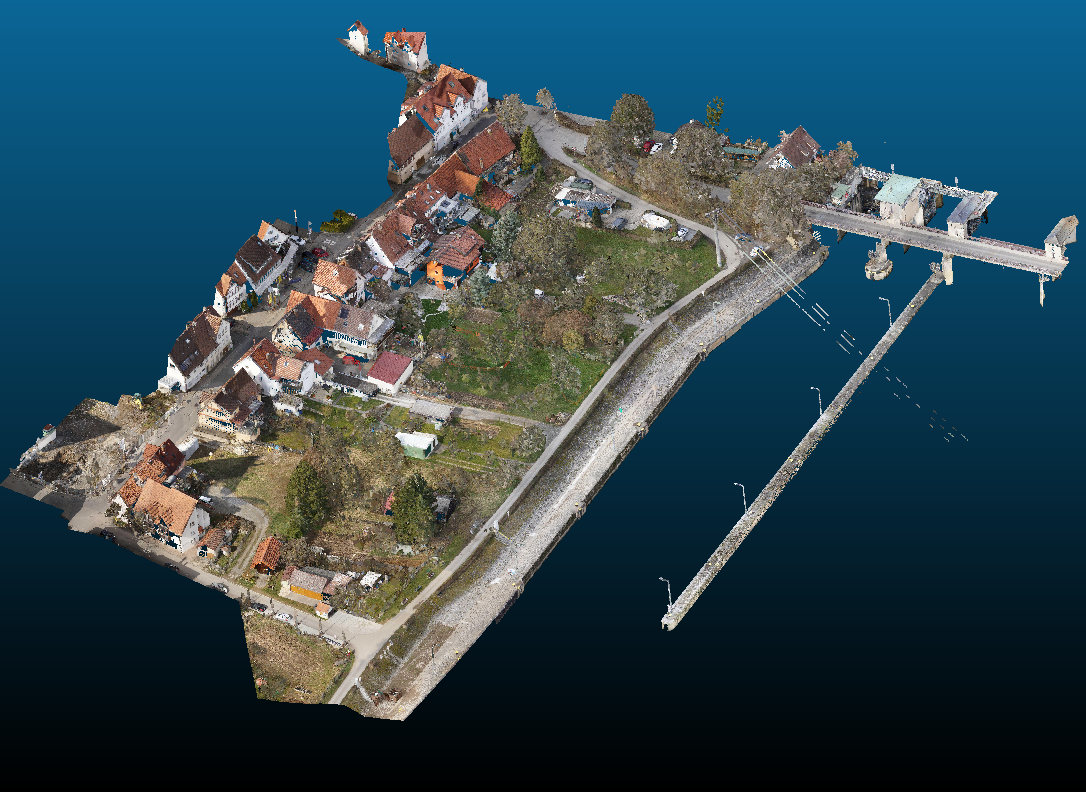}
        \includegraphics[width=\textwidth, trim=0cm 3cm 0cm 0cm, clip]{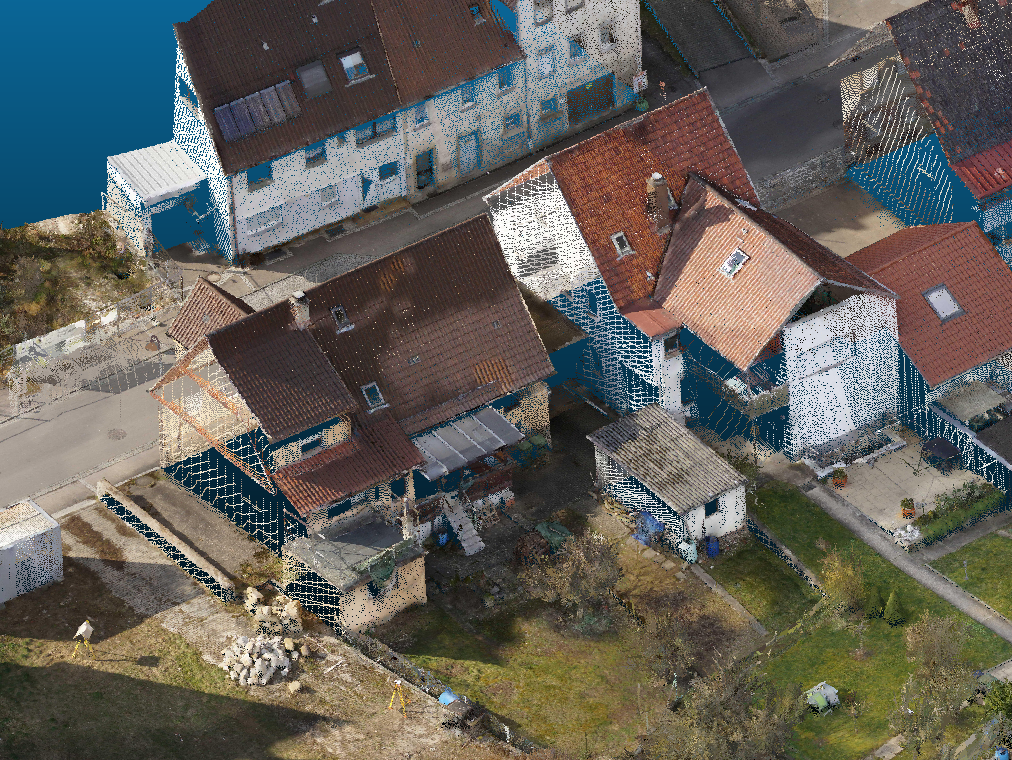}
        \caption{RGB}
        \label{fig:hessigheim_rgb}
    \end{subfigure}
    \begin{subfigure}{0.3\textwidth}
        \centering
        \includegraphics[width=\textwidth]{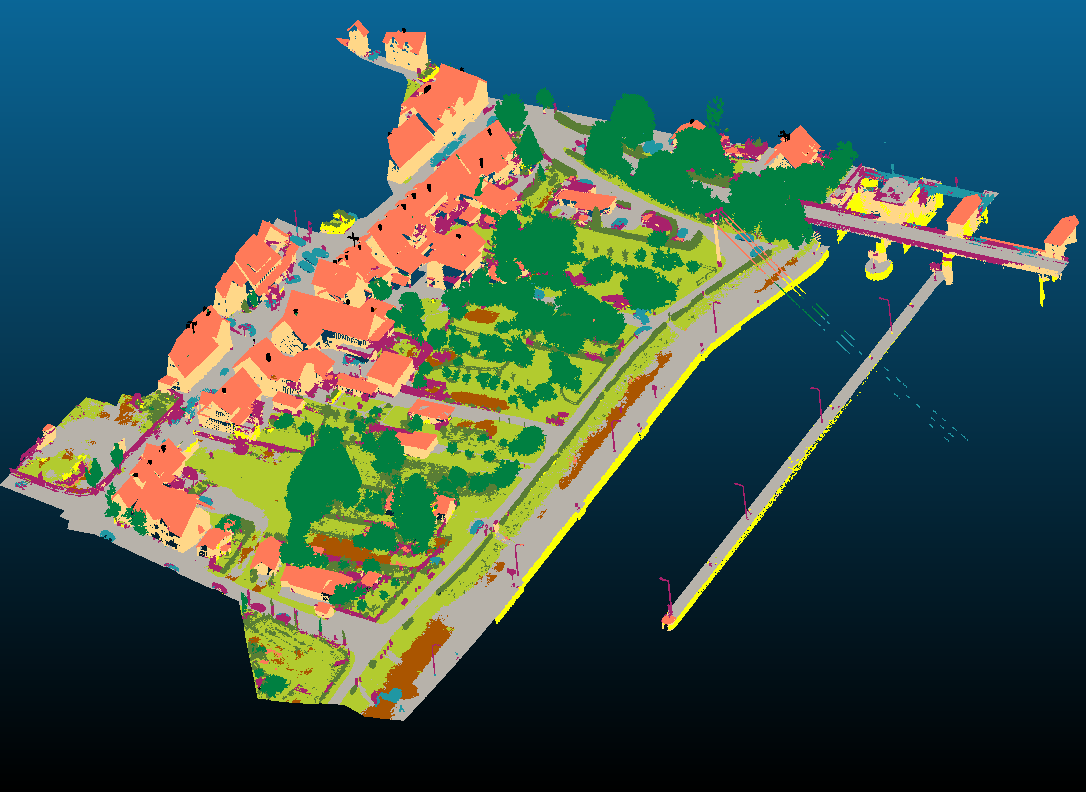}
        \includegraphics[width=\textwidth, trim=0cm 3cm 0cm 0cm, clip]{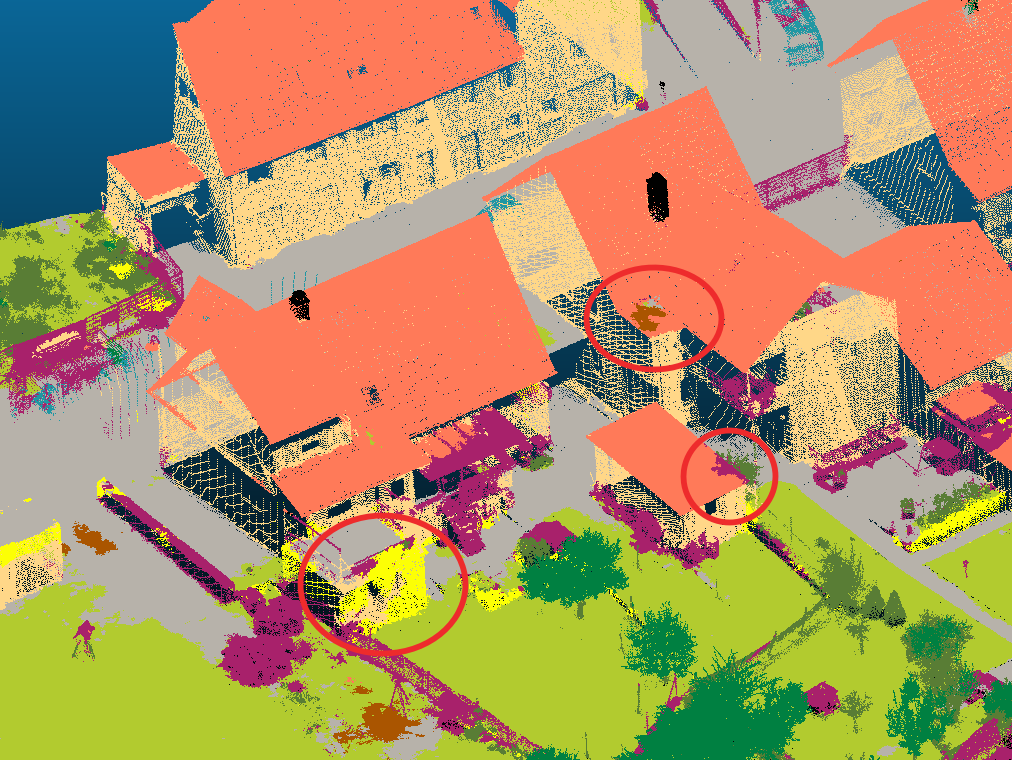}
        \caption{Pred. ``color''}
        \label{fig:hessigheim_color}
    \end{subfigure}
    \begin{subfigure}{0.3\textwidth}
        \centering
        \includegraphics[width=\textwidth]{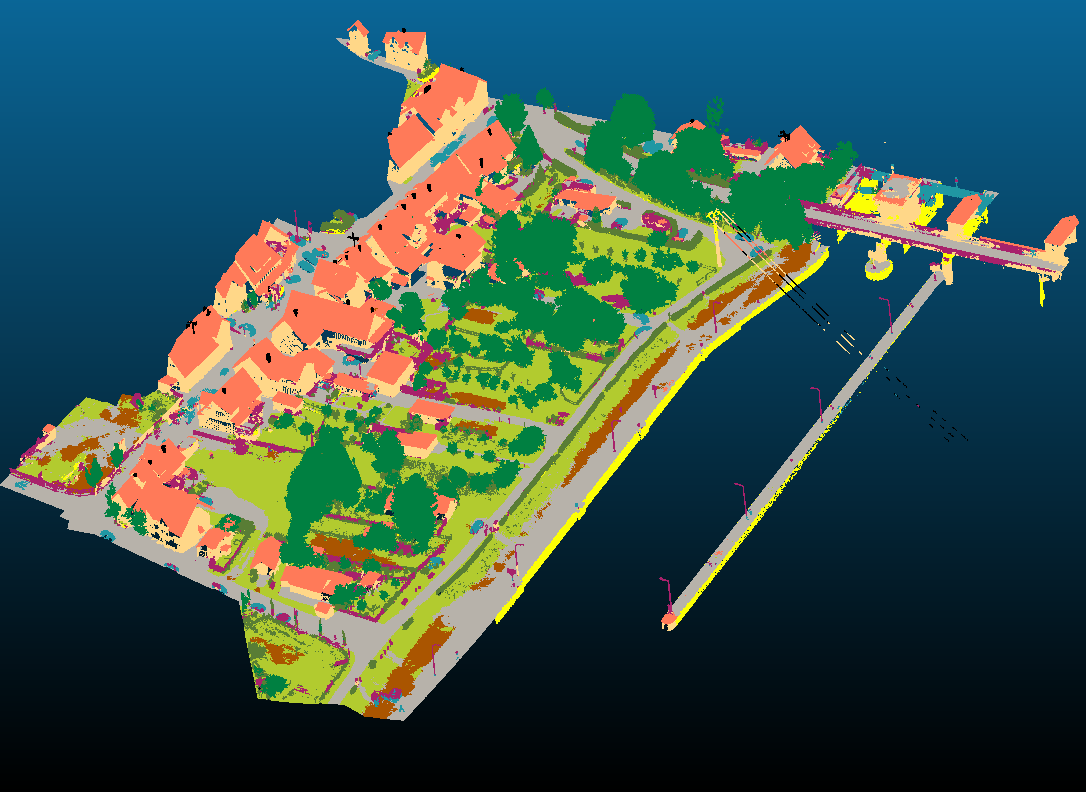}
        \includegraphics[width=\textwidth, trim=0cm 3cm 0cm 0cm, clip]{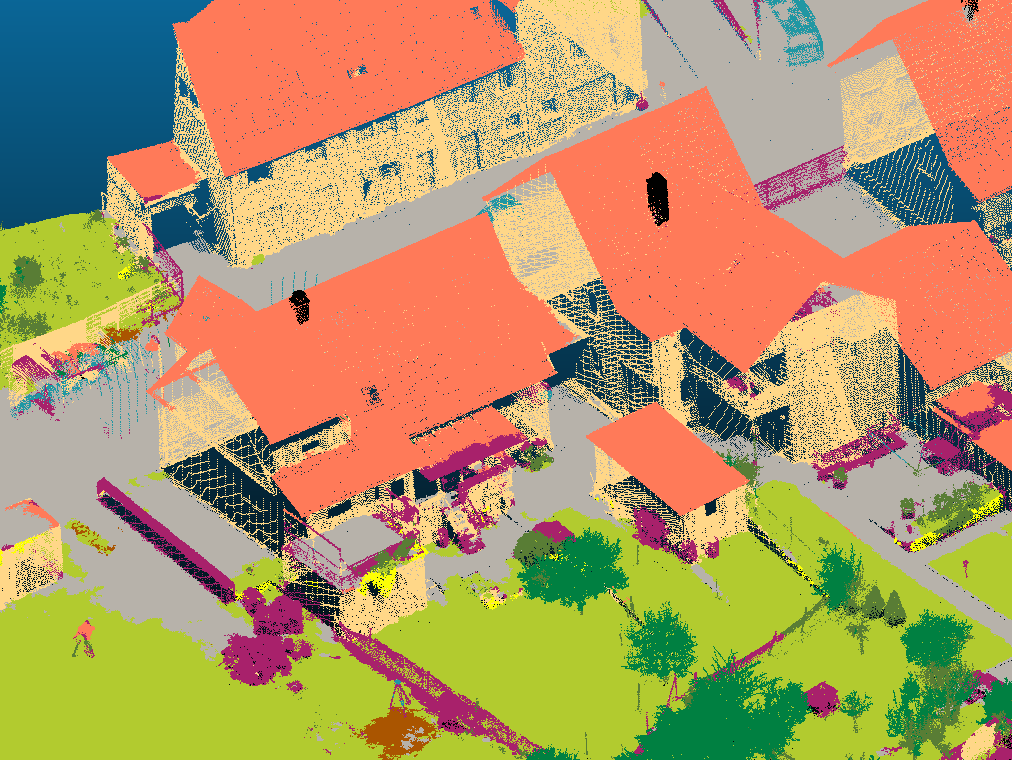}
        \caption{Pred. ``color+$h_r$''}
        \label{fig:hessigheim_color_relz}
    \end{subfigure}
    \caption{The test set of the Hessigheim March 2018 capture and predictions with different configurations in its entirety, and a zoomed in area. \textcolor{newcolor}{Errors in Fig.~\ref{fig:hessigheim_color} are circled in red.} Ground truth is not publicly available.}
    \label{fig:hessigheim}
\end{figure*}

As shown in Table~\ref{tab:hessigheim}, three variations on the Hessigheim dataset were submitted and evaluated. First, to establish the baseline, a model with just color was trained, reaching an average F1 score of 72.35\%. Adding relative elevation, the average F1 score improves substantially by almost 4 \textcolor{cameraready}{pp.} to 76.01\%.
The largest improvements with relative elevation are the shrub, gravel, and vertical surface class. All of these classes benefit from the elevation information, shrubs for example can be better distinguished from the ground and tree. 
Most other classes improve as well. It should be noted that the improvement of large classes, like of buildings, trees, and the ground, in the Hessigheim dataset as well as in the Swiss3DCities dataset, will be less pronounced. 
This is because the improvements mainly affect the challenging ambiguous points, which represent a small fraction in large classes and thus have limited influence on the overall score.
Adding \textcolor{newcolor}{both elevation and local} features, however, degrades the performance of the network, but it is still better \textcolor{cameraready}{in terms of mF1} than the color only configuration. 
This is attributed to the fact that the point density, especially on vertical surfaces, is very heterogeneous in the Hessigheim dataset due to more down facing LiDAR scan sweeps, whereas the photogrammetric Swiss3DCities dataset has a quite homogeneous point cloud. 
This makes features like $\nu$ or $\sigma(z)$ inconsistent. Furthermore, hand-crafted features require setting parameters like the number of neighbors, \textcolor{cameraready}{for which sensible choices} may differ between datasets.

Figure~\ref{fig:hessigheim} shows the qualitative results of the baseline ``color'' and ``color+$h_r$'' configuration, along with the RGB input point cloud. Since this is the test set, a ground truth cannot be shown. The detection of roofs (red) with relative elevation in Figure~\ref{fig:hessigheim_color_relz} is slightly improved, showing less confusion with the urban furniture class (purple). Facades (orange) are also less confused with the vertical surface class (yellow).
The uneven point density is apparent in the figures, where individual LiDAR sweeps can be seen on the vertical surfaces as well as occluded areas like underneath protruding roofs.

\begin{table*}[ht]
\caption{Performance of RandLA-Net on the Toronto 3D L002 validation tile.} 
\label{tab:toronto3d}
\resizebox{\textwidth}{!}{%
\begin{tabular}{llllllllllll}
\hline
\textbf{Feat.} &
  \textbf{OA} &
  \textbf{\begin{tabular}[c]{@{}l@{}}Mean\\ IoU\end{tabular}} &
  \textbf{\begin{tabular}[c]{@{}l@{}}IoU\\ uncl.\end{tabular}} &
  \textbf{\begin{tabular}[c]{@{}l@{}}IoU\\ road\end{tabular}} &
  \textbf{\begin{tabular}[c]{@{}l@{}}IoU\\ r. ma.\end{tabular}} &
  \textbf{\begin{tabular}[c]{@{}l@{}}IoU\\ natu.\end{tabular}} &
  \textbf{\begin{tabular}[c]{@{}l@{}}IoU\\ build.\end{tabular}} &
  \textbf{\begin{tabular}[c]{@{}l@{}}IoU\\ ut. l.\end{tabular}} &
  \textbf{\begin{tabular}[c]{@{}l@{}}IoU\\ pole\end{tabular}} &
  \textbf{\begin{tabular}[c]{@{}l@{}}IoU\\ car\end{tabular}} &
  \textbf{\begin{tabular}[c]{@{}l@{}}IoU\\ fence\end{tabular}} \\ \hline
color &
  92.80 &
  71.52 &
  41.47 &
  92.30 &
  56.23 &
  91.95 &
  88.67 &
  83.88 &
  \textbf{78.37} &
  \textbf{87.97} &
  \textbf{22.84} \\
color+$h_r$ &
  \textbf{93.43} &
  \textbf{72.10} &
  \textbf{44.52} &
  \textbf{93.35} &
  \textbf{62.32} &
  \textbf{92.67} &
  \textbf{89.81} &
  \textbf{84.84} &
  73.73 &
  86.10 &
  21.60 \\
all &
  90.26 &
  61.66 &
  19.58 &
  90.11 &
  17.78 &
  89.25 &
  81.17 &
  81.01 &
  69.44 &
  84.79 &
  21.78 \\\hline
\end{tabular}%
}
\end{table*}

\begin{figure*}[h]
    \centering
    \begin{subfigure}{0.45\textwidth}
        \centering
        \includegraphics[width=\textwidth, trim=0cm 0cm 0cm 2cm, clip]{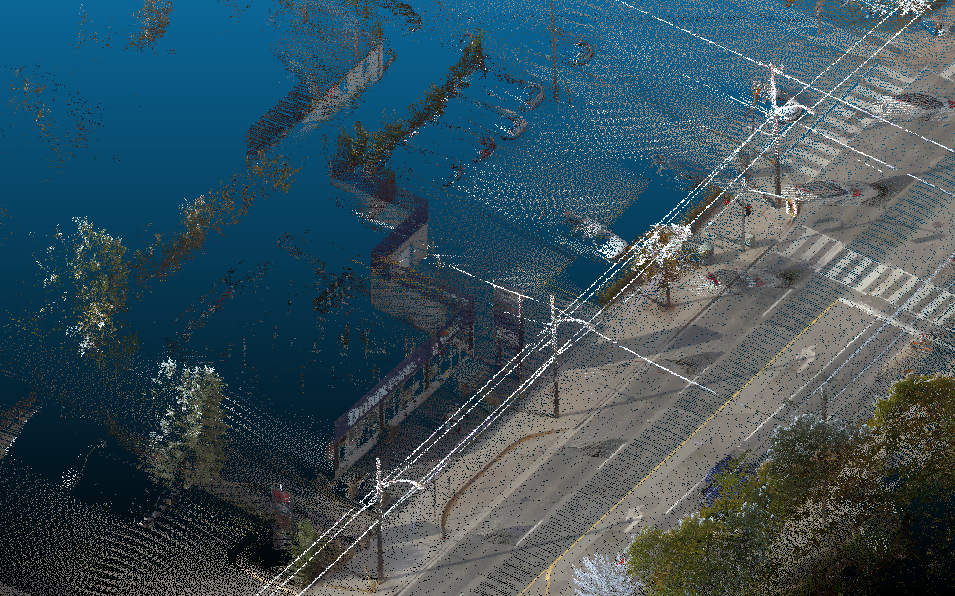}
        \caption{RGB}
        \label{fig:toronto_rgb}    
    \end{subfigure}
    \hfill
    \begin{subfigure}{0.45\textwidth}
        \centering
        \includegraphics[width=\textwidth, trim=0cm 0cm 0cm 2cm, clip]{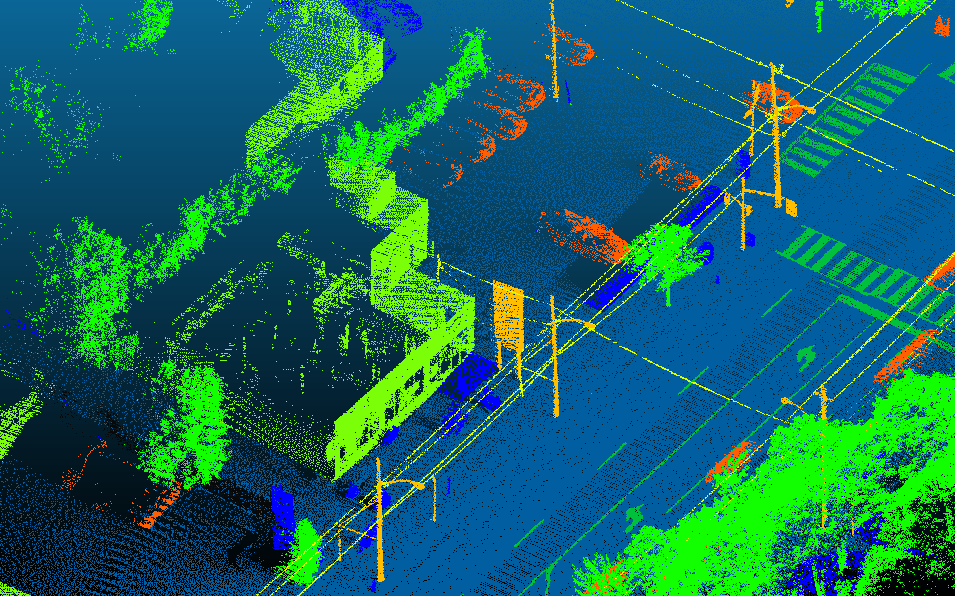}
        \caption{Ground truth}
        \label{fig:toronto_gt}
    \end{subfigure}
    \begin{subfigure}{0.45\textwidth}
        \vspace{0.25cm}
        \centering
        \includegraphics[width=\textwidth, trim=0cm 0cm 0cm 2cm, clip]{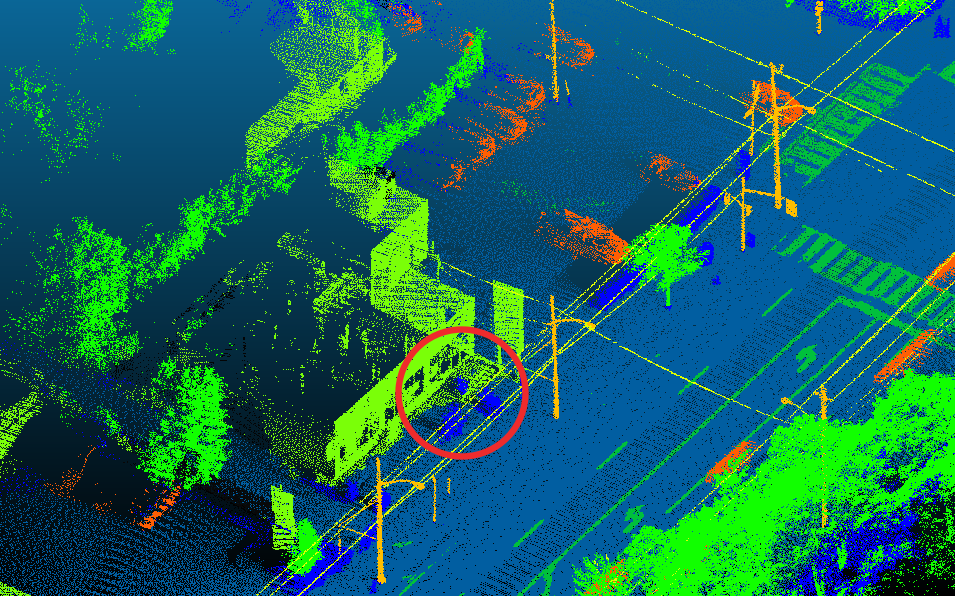}
        \caption{Prediction ``color''}
        \label{fig:toronto_color}
        \vspace{0.25cm}
    \end{subfigure}
    \hfill
    \begin{subfigure}{0.45\textwidth}
        \vspace{0.25cm}
        \centering
        \includegraphics[width=\textwidth, trim=0cm 0cm 0cm 2cm, clip]{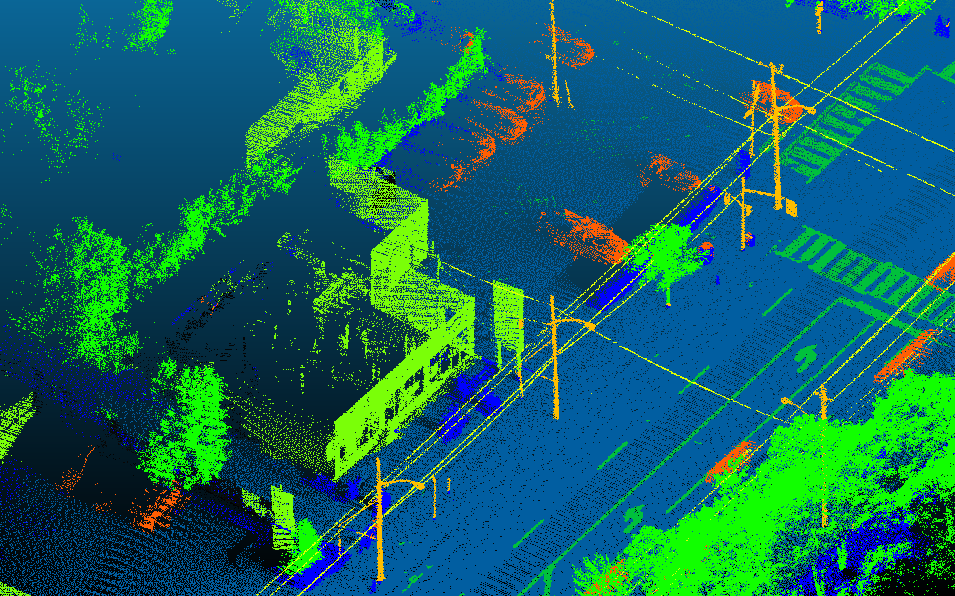}
        \caption{Prediction ``color+$h_r$''}
        \label{fig:toronto_color_relz}
        \vspace{0.25cm}
    \end{subfigure}
    \caption{A zoomed in area of the Toronto3D L002 validation tile with the predictions of two configurations and ground truth. The tile has been downsampled using grid sampling. \textcolor{newcolor}{An error in Fig.~\ref{fig:toronto_color} is circled in red.}}
    \label{fig:toronto3d}
\end{figure*}

On the last dataset, Toronto3D, relative elevation brings a small improvement, as seen in Table~\ref{tab:toronto3d}, where the mIoU jumps over 0.5 pp. to 72.10\%. The small improvement could be explained by the relative elevation not being fully accurate due to the RANSAC approximation. 
The addition of \textcolor{newcolor}{local} features, however, severely degrades the performance, making it worse than the color configuration. One reasonable explanation is that the features are calculated on the full point cloud, where the point density varies wildly, while for RandLA-Net the points and features are first grid sampled (grid size \SI{6}{\centi\meter}, just like the original authors of RandLA-Net did). The downsampling also increases the receptive field in terms of metric distance, since the point density is drastically reduced, making $h_r$ less impactful.
No downsampling was applied to the other datasets for training or evaluation. Another reason is that the strong density variation and occlusion effects \textcolor{newcolor}{make} less useful features, especially the 2D features like $\nu$ or $\sigma(z)$. Density variations are also caused by the distance from the vehicle and vehicle speed, both of which are not correlated to useful information for semantic segmentation. 
Note that the results of Toronto3D on vanilla RandLA-Net do not reach the mIoU of the original authors. 
This may be due to weaker hardware or a different implementation of RandLA-Net. 
The RGB point cloud, ground truth and qualitative results are shown in Fig.~\ref{fig:toronto3d}. Small improvements can be seen in the delineation of road and ground. The point density is, of course, highest along the road, where the vehicle was driven.

\section{Conclusion and \textcolor{newcolor}{O}utlook}
With 3D outdoor point clouds, it is relatively straightforward to compute a Digital Terrain Model, or DTM, using well established methods. Then, the relative elevation of individual points of the point cloud with respect to the ground surface can be determined.
A consistent and large improvement in semantic segmentation accuracy of aerial point clouds is achieved by integrating this relative elevation information into an existing deep learning network, such as  RandLA-Net, \textcolor{newcolor}{without needing to modify its architecture}. \textcolor{newcolor}{This approach renders the source or method of DTM derivation irrelevant. It could utilize traditional numerical methods, DL techniques, or even DTMs from publicly available government data sources.}
The addition of the relative elevation feature ensures ground awareness through a long-range dependency between ground and object points, especially in cases where the receptive field does not capture the terrain.
\textcolor{cameraready}{In the Hessigheim dataset, the addition of the elevation information increases the mF1 from 72.35\% to 76.01\%, and in the Swiss3DCities dataset the mIoU from 66.70\% to 69.81\%.}
Other \textcolor{newcolor}{local features} 
 are less reliably beneficial. \textcolor{cameraready}{Using these local features can even degrade network performance in more datasets with more inhomogeneous point distributions, like in both LiDAR datasets of Hessigheim and Toronto3D.
There are two main reasons for this, which contrast the relative elevation:  1) \textcolor{cameraready}{The local features} depend on the structure of the point cloud 2) they are quite localized and easily captured by the receptive field of the network.}

It is therefore advisable to incorporate relative elevation information into classification and segmentation tasks in \textcolor{kevincolor}{large outdoor point clouds, which are typical for} remote sensing, whenever feasible. This recommendation holds not only for 2D scenarios, as demonstrated in~\cite{qiu2022exploring} and~\cite{audebert2017rgb}, but also for 3D point cloud segmentation tasks using modern deep learning networks, as evidenced by the findings presented in this paper\textcolor{newcolor}{, even though the input point cloud already implicitly contains the information necessary to derive the DTM}. \textcolor{cameraready}{The effect of relative elevation could be even more apparent in datasets with larger or taller buildings and higher resolution.}

\textcolor{dimicolor}{The prevailing direction in Computer Vision involves shifting towards methods that require fewer or even no annotated labels at all, as the process of labeling is often prohibitively expensive. In future work, we therefore may integrate} elevation features into other point cloud segmentation models and even explore their effect on self-supervision tasks and in transformer models. 
\vspace{-4mm}
\bibliographystyle{apalike}
\bibliography{egbib}
\end{document}